\documentclass[11pt,letterpaper]{article}
\usepackage{emnlp2017}
\usepackage{times}
\usepackage{latexsym}
\usepackage{multirow}
\usepackage{enumitem}
\usepackage{caption}
\usepackage{subcaption}
\usepackage{graphicx}
\usepackage{booktabs}
\usepackage{makecell}

\newenvironment{querytermlist}{\begin{itemize}[noitemsep]}{\end{itemize}}

\emnlpfinalcopy

\def\emnlppaperid{***}

\title{\sc{Quasar}: Datasets for Question Answering by\\Search and Reading}

\author{Bhuwan Dhingra \qquad Kathryn Mazaitis \qquad William W. Cohen \\
		School of Computer Science \\
        Carnegie Mellon University \\
  {\tt \{bdhingra, krivard, wcohen\}@cs.cmu.edu}}

\date{}

\begin{document}

\maketitle

\begin{abstract}
We present two new large-scale datasets aimed at evaluating systems designed to comprehend a natural language query and extract its answer from a large corpus of text.
The \textsc{Quasar-S} dataset consists of 37000 cloze-style (fill-in-the-gap) queries constructed from definitions of software entity tags on the popular website Stack Overflow. The posts and comments on the website serve as the background corpus for answering the cloze questions.
The \textsc{Quasar-T} dataset consists of 43000 open-domain trivia questions and their answers obtained from various internet sources. ClueWeb09 \citep{callan2009clueweb09} serves as the background corpus for extracting these answers.
We pose these datasets as a challenge for two related subtasks of factoid Question Answering: (1) searching for relevant pieces of text that include the correct answer to a query, and (2) reading the retrieved text to answer the query. We also describe a retrieval system for extracting relevant sentences and documents from the corpus given a query, and include these in the release for researchers wishing to only focus on (2).
We evaluate several baselines on both datasets, ranging from simple heuristics to powerful neural models, and show that these lag behind human performance by $16.4\%$ and $32.1\%$ for \textsc{Quasar-S} and \textsc{-T} respectively. The datasets are available at \url{https://github.com/bdhingra/quasar}.
\end{abstract}

\section{Introduction}
% Question Answering (QA) systems aim to directly answer user queries rather than retrieve a ranked list of documents which contain the answer. In particular, 
Factoid Question Answering (QA) aims to extract answers, from an underlying knowledge source, to information seeking questions posed in natural language. Depending on the knowledge source available there are two main approaches for factoid QA. Structured sources, including Knowledge Bases (KBs) such as Freebase \citep{bollacker2008freebase}, are easier to process automatically since the information is organized according to a fixed schema. In this case the question is parsed into a logical form in order to query against the KB. However, even the largest KBs are often incomplete \citep{miller2016key,west2014knowledge}, and hence can only answer a limited subset of all possible factoid questions.

For this reason the focus is now shifting towards unstructured sources, such as Wikipedia articles, which hold a vast quantity of information in textual form and, in principle, can be used to answer a much larger collection of questions. Extracting the correct answer from unstructured text is, however, challenging, and typical QA pipelines consist of the following two components: (1) \textit{searching} for the passages relevant to the given question, and (2) \textit{reading} the retrieved text in order to select a span of text which best answers the question \citep{chen2017reading,watanabe2017question}.

Like most other language technologies, the current research focus for both these steps is firmly on machine learning based approaches for which performance improves with the amount of data available. Machine reading performance, in particular, has been significantly boosted in the last few years with the introduction of large-scale reading comprehension datasets such as CNN / DailyMail \citep{hermann2015teaching} and Squad \citep{rajpurkar2016squad}. State-of-the-art systems for these datasets \citep{dhingra2016gated,seo2016bidirectional} focus solely on step (2) above, in effect assuming the relevant passage of text is already known. 

\begin{figure*}[!htbp]
\small
\begin{tabular}{rp{0.8\textwidth}}
\textbf{Question} & javascript -- javascript not to be confused with java is a dynamic weakly-typed language used for XXXXX as well as server-side scripting . \\
\textbf{Answer} & \textbf{client-side} \\
\textbf{Context excerpt} & JavaScript is not weakly typed, it is strong typed. \newline
JavaScript is a \textbf{Client Side} Scripting Language. \newline
JavaScript was the **original** \textbf{client-side} web scripting language.\\\\
\textbf{Question} & 7-Eleven stores were temporarily converted into Kwik E-marts to promote the release of what movie? \\
\textbf{Answer} & \textbf{the simpsons movie} \\
\textbf{Context excerpt} & In July 2007 , 7-Eleven redesigned some stores to look like Kwik-E-Marts in select cities to promote \textbf{The Simpsons Movie} . \newline
Tie-in promotions were made with several companies , including 7-Eleven , which transformed selected stores into Kwik-E-Marts . \newline
`` 7-Eleven Becomes Kwik-E-Mart for ` \textbf{Simpsons Movie} ' Promotion '' .
\end{tabular}
\caption{\small Example short-document instances from \textsc{Quasar-S} (top) and \textsc{Quasar-T} (bottom)}\label{f_demo}
\end{figure*}

In this paper, we introduce two new datasets for QUestion Answering by Search And Reading -- \textsc{Quasar}. The datasets each consist of factoid question-answer pairs and a corresponding large background corpus to facilitate research into the combined problem of retrieval and comprehension. \textsc{Quasar-S} consists of 37,362 cloze-style questions constructed from definitions of software entities available on the popular website Stack Overflow\footnote{Stack Overflow is a website featuring questions and answers (posts) from a wide range of topics in computer programming. The entity definitions were scraped from \url{https://stackoverflow.com/tags}.}. The answer to each question is restricted to be another software entity, from an output vocabulary of 4874 entities.
\textsc{Quasar-T} consists of 43,013 trivia questions collected from various internet sources by a trivia enthusiast. The answers to these questions are free-form spans of text, though most are noun phrases.

While production quality QA systems may have access to the entire world wide web as a knowledge source, for \textsc{Quasar} we restrict our search to specific background corpora. This is necessary to avoid uninteresting solutions which directly extract answers from the sources from which the questions were constructed. For \textsc{Quasar-S} we construct the knowledge source by collecting top 50 threads\footnote{A question along with the answers provided by other users is collectively called a thread. The threads are ranked in terms of votes from the community. Note that these questions are different from the cloze-style queries in the \textsc{Quasar-S} dataset.} tagged with each entity in the dataset on the Stack Overflow website. For \textsc{Quasar-T} we use ClueWeb09 \citep{callan2009clueweb09}, which contains about 1 billion web pages collected between January and February 2009. Figure~\ref{f_demo} shows some examples.

Unlike existing reading comprehension tasks, the \textsc{Quasar} tasks go beyond the ability to only understand a given passage, and require the ability to answer questions given large corpora. Prior datasets (such as those used in \citep{chen2017reading}) are constructed by first selecting a passage and then constructing questions about that passage. This design (intentionally) ignores some of the subproblems required to answer open-domain questions from corpora, namely searching for passages that may contain candidate answers, and aggregating information/resolving conflicts between candidates from many passages. The purpose of Quasar is to allow research into these subproblems, and in particular whether the search step can benefit from integration and joint training with downstream reading systems. 

Additionally, \textsc{Quasar-S} has the interesting feature of being a closed-domain dataset about computer programming, and successful approaches to it must develop domain-expertise and a deep understanding of the background corpus. To our knowledge it is one of the largest closed-domain QA datasets available.
%, and hence can be used for research into data-intensive deep learning expert systems. 
\textsc{Quasar-T}, on the other hand, consists of open-domain questions based on trivia, which refers to ``bits of information, often of little importance".
Unlike previous open-domain systems which rely heavily on the redundancy of information on the web to correctly answer questions,
%Trivia, however, refers to 
% \footnote{\url{https://en.wikipedia.org/wiki/Trivia}}
we hypothesize that \textsc{Quasar-T} requires a deeper reading of documents to answer correctly. 
% \textsc{QUASAR-T} is an open-domain QA dataset since its questions span a wide variety of topics ranging from movies to history and religion. 
% In the short-term, research can focus on one or the other of these datasets to either build a general purpose or a domain expert QA system. A longer-term goal might be to build a common system which can perform well on both datasets.

We evaluate \textsc{Quasar} against human testers, as well as several baselines ranging from na{\"i}ve heuristics to state-of-the-art machine readers. The best performing baselines achieve $33.6\%$ and $28.5\%$ on \textsc{Quasar-S} and \textsc{Quasar-T}, while human performance is $50\%$ and $60.6\%$ respectively. For the automatic systems, we see an interesting tension between searching and reading accuracies -- retrieving more documents in the search phase leads to a higher coverage of answers, but makes the comprehension task more difficult. We also collect annotations on a subset of the development set questions to allow researchers to analyze the categories in which their system performs well or falls short. We plan to release these annotations along with the datasets, and our retrieved documents for each question.

\section{Existing Datasets}
% Next we discuss existing datasets for Question Answering and how \textsc{Quasar} relates to them.

\paragraph{Open-Domain QA:} Early research into open-domain QA was driven by the TREC-QA challenges organized by the National Institute of Standards and Technology (NIST) \citep{voorhees2000building}. Both dataset construction and evaluation were done manually, restricting the size of the dataset to only a few hundreds. \textsc{WikiQA} \citep{yang2015wikiqa} was introduced as a larger-scale dataset for the subtask of answer sentence selection, however it does not identify spans of the actual answer within the selected sentence. More recently, \citet{miller2016key} introduced the \textsc{MoviesQA} dataset where the task is to answer questions about movies from a background corpus of Wikipedia articles. \textsc{MoviesQA} contains $\sim100k$ questions, however many of these are similarly phrased and fall into one of only $13$ different categories; hence, existing systems already have $\sim85\%$ accuracy on it \citep{watanabe2017question}. MS MARCO \citep{nguyen2016ms} consists of diverse real-world queries collected from Bing search logs, however many of them not factual, which makes their evaluation tricky. \citet{chen2017reading} study the task of \textit{Machine Reading at Scale} which combines the aspects of search and reading for open-domain QA. They show that jointly training a neural reader on several distantly supervised QA datasets leads to a performance improvement on all of them. This justifies our motivation of introducing two new datasets to add to the collection of existing ones; more data is good data.

\paragraph{Reading Comprehension:} Reading Comprehension (RC) aims to measure the capability of systems to ``understand'' a given piece of text, by posing questions over it. It is assumed that the passage containing the answer is known beforehand. Several datasets have been proposed to measure this capability. \citet{richardson2013mctest} used crowd-sourcing to collect MCTest -- $500$ stories with $2000$ questions over them. Significant progress, however, was enabled when \citet{hermann2015teaching} introduced the much larger CNN / Daily Mail datasets consisting of $300k$ and $800k$ cloze-style questions respectively. Children's Book Test (CBT) \citep{hill2015goldilocks} and Who-Did-What (WDW) \citep{onishi2016did} are similar cloze-style datasets. However, the automatic procedure used to construct these questions often introduces ambiguity and makes the task more difficult \citep{chen2016thorough}. Squad \citep{rajpurkar2016squad} and NewsQA \citep{trischler2016newsqa} attempt to move toward more general extractive QA by collecting, through crowd-sourcing, more than $100k$ questions whose answers are spans of text in a given passage. Squad in particular has attracted considerable interest, but recent work \citep{weissenborn2017fastqa} suggests that answering the questions does not require a great deal of reasoning. 

Recently, \citet{joshi2017triviaqa} prepared the TriviaQA dataset, which also consists of trivia questions collected from online sources, and is similar to \textsc{Quasar-T}. 
% Since TriviaQA was not publicly available at the time of writing, it is unclear how much, if any, overlap exists between the two. Regardless, o
However, the documents retrieved for TriviaQA were obtained using a commercial search engine, making it difficult for researchers to vary the retrieval step of the QA system in a controlled fashion; in contrast we use ClueWeb09, a standard corpus.  We also supply a larger collection of retrieved passages, including many not containing the correct answer to facilitate research into retrieval, perform a more extensive analysis of baselines for answering the questions, and provide additional human evaluation and annotation of the questions. In addition we present \textsc{Quasar-S}, a second dataset.
% The LAMBADA datset \citep{paperno2016lambada} is another interesting language comprehension dataset which has been shown to involve aspects of reading comprehension \citep{chu2016broad}. 
SearchQA \citep{dunn2017searchqa} is another recent dataset aimed at facilitating research towards an end-to-end QA pipeline, however this too uses a commercial search engine, and does not provide negative contexts not containing the answer, making research into the retrieval component difficult.

\section{Dataset Construction}

Each dataset consists of a collection of records with one QA problem per record. For each record, we include some question text, a context document relevant to the question, a set of candidate solutions, and the correct solution. In this section, we describe how each of these fields was generated for each \textsc{Quasar} variant.

\subsection{Question sets}

\paragraph{\textsc{Quasar-S}:} The software question set was built from the definitional ``excerpt'' entry for each tag (entity) on 
% the software development advice site 
StackOverflow. For example the excerpt for the ``java`` tag is, ``Java is a general-purpose object-oriented programming language designed to be used in conjunction with the Java Virtual Machine (JVM).'' Not every excerpt includes the tag being defined (which we will call the ``head tag''), so we prepend the head tag to the front of the string to guarantee relevant results later on in the pipeline. We then completed preprocessing of the software questions by downcasing and tokenizing the string using a custom tokenizer compatible with special characters in software terms (e.g. ``.net'', ``c++''). Each preprocessed excerpt was then converted to a series of cloze questions using a simple heuristic: first searching the string for mentions of other entities, then repleacing each mention in turn with a placeholder string (Figure \ref{f_clozeExamples}).

\begin{figure*}[!htbp]
\small
\begin{tabular}{cp{0.8\textwidth}}
%\multicolumn{2}{c}{} \\\hline
\textbf{Excerpt} & Java is a general-purpose object-oriented programming language designed to be used in conjunction with the Java Virtual Machine (JVM). \\
%\multicolumn{2}{c}{} \\\hline
\makecell{\textbf{Preprocessed} \\ \textbf{Excerpt}} & java --- java is a general-purpose object-oriented programming language designed to be used in conjunction with the java virtual-machine jvm .\\\\
\multicolumn{2}{c}{\textbf{Cloze Questions}} \\
\textit{Cloze} & \textit{Question} \\\hline
java & java --- java is a general-purpose object-oriented programming language designed to be used in conjunction with the @placeholder virtual-machine jvm . \\\hline
virtual-machine & java --- java is a general-purpose object-oriented programming language designed to be used in conjunction with the java @placeholder jvm . \\\hline
jvm & java --- java is a general-purpose object-oriented programming language designed to be used in conjunction with the java virtual-machine @placeholder . \\
\end{tabular}
\caption{\small Cloze generation}\label{f_clozeExamples}
\end{figure*}

This heuristic is noisy, since the software domain often overloads existing English words (e.g. ``can'' may refer to a Controller Area Network bus; ``swap'' may refer to the temporary storage of inactive pages of memory on disk; ``using'' may refer to a namespacing keyword). To improve precision we scored each cloze based on the relative incidence of the term in an English corpus versus in our StackOverflow one, and discarded all clozes scoring below a threshold. This means our dataset does not include any cloze questions for terms which are common in English (such as ``can'' ``swap'' and ``using'', but also ``image'' ``service'' and ``packet''). A more sophisticated entity recognition system could make recall improvements here.

\paragraph{\textsc{Quasar-T}:} The trivia question set was built from a collection of just under 54,000 trivia questions collected by Reddit user 007craft and released in December 2015\footnote{\url{https://www.reddit.com/r/trivia/comments/3wzpvt/free_database_of_50000_trivia_questions/}}. The raw dataset was noisy, having been scraped from multiple sources with variable attention to detail in formatting, spelling, and accuracy. We filtered the raw questions to remove unparseable entries as well as any True/False or multiple choice questions, for a total of ~52,000 free-response style questions remaining. The questions range in difficulty, from straightforward (``Who recorded the song `Rocket Man''' ``Elton John'') to difficult (``What was Robin Williams paid for Disney's Aladdin in 1982'' ``Scale \$485 day + Picasso Painting'') to debatable (``According to Earth Medicine what's the birth totem for march'' ``The Falcon'')\footnote{In Earth Medicine, March has two birth totems, the falcon and the wolf.}

\subsection{Context Retrieval}
\label{sec:retreival}
The context document for each record consists of a list of ranked and scored pseudodocuments relevant to the question. 

Context documents for each query were generated in a two-phase fashion, first collecting a large pool of semirelevant text, then filling a temporary index with short or long pseudodocuments from the pool, and finally selecting a set of $N$ top-ranking pseudodocuments (100 short or 20 long) from the temporary index.

For \textsc{Quasar-S}, the pool of text for each question was composed of 50+ question-and-answer threads scraped from \url{http://stackoverflow.com}. StackOverflow keeps a running tally of the top-voted questions for each tag in their knowledge base; we used Scrapy\footnote{\url{https://scrapy.org}} to pull the top 50 question posts for each tag, along with any answer-post responses and metadata (tags, authorship, comments). From each thread we pulled all text not marked as code, and split it into sentences using the Stanford NLP sentence segmenter, truncating sentences to 2048 characters. Each sentence was marked with a thread identifier, a post identifier, and the tags for the thread. Long pseudodocuments were either the full post (in the case of question posts), or the full post and its head question (in the case of answer posts), comments included. Short pseudodocuments were individual sentences.

To build the context documents for \textsc{Quasar-S}, the pseudodocuments for the entire corpus were loaded into a disk-based lucene index, each annotated with its thread ID and the tags for the thread. This index was queried for each cloze using the following lucene syntax:

\begin{querytermlist}
\small
\item[] SHOULD(PHRASE(question text))
\item[] SHOULD(BOOLEAN(question text))
\item[] MUST(tags:\$headtag)
\end{querytermlist}

where ``question text'' refers to the sequence of tokens in the cloze question, with the placeholder removed. The first SHOULD term indicates that an exact phrase match to the question text should score highly. The second SHOULD term indicates that any partial match to tokens in the question text should also score highly, roughly in proportion to the number of terms matched. The MUST term indicates that only pseudodocuments annotated with the head tag of the cloze should be considered.

The top $100N$ pseudodocuments were retrieved, and the top $N$ unique pseudodocuments were added to the context document along with their lucene retrieval score.  Any questions showing zero results for this query were discarded.

For \textsc{Quasar-T}, the pool of text for each question was composed of 100 HTML documents retrieved from ClueWeb09. Each question-answer pair was converted to a {\tt\#combine} query in the Indri query language to comply with the ClueWeb09 batch query service, using simple regular expression substitution rules to remove ({\verb~s/[.(){}<>:*`_]+//g~}) or replace ({\verb~s/[,?']+/ /g~}) illegal characters. Any questions generating syntax errors after this step were discarded. We then extracted the plaintext from each HTML document using Jericho\footnote{\url{http://jericho.htmlparser.net/docs/index.html}}. For long pseudodocuments we used the full page text, truncated to 2048 characters. For short pseudodocuments we used individual sentences as extracted by the Stanford NLP sentence segmenter, truncated to 200 characters.

To build the context documents for the trivia set, the pseudodocuments from the pool were collected into an in-memory lucene index and queried using the question text only (the answer text was not included for this step). The structure of the query was identical to the query for \textsc{Quasar-S}, without the head tag filter:

\begin{querytermlist}
\small
\item[] SHOULD(PHRASE(question text)) 
\item[] SHOULD(BOOLEAN(question text))
\end{querytermlist}
  
The top $100N$ pseudodocuments were retrieved, and the top $N$ unique pseudodocuments were added to the context document along with their lucene retrieval score. Any questions showing zero results for this query were discarded.

\subsection{Candidate solutions}
The list of candidate solutions provided with each record is guaranteed to contain the correct answer to the question. 
\textsc{Quasar-S} used a closed vocabulary of 4874 tags as its candidate list.
Since the questions in \textsc{Quasar-T} are in free-response format, we constructed a separate list of candidate solutions for each question. Since most of the correct answers were noun phrases, we took each sequence of {\tt NN*} -tagged tokens in the context document, as identified by the Stanford NLP Maxent POS tagger, as the candidate list for each record. If this list did not include the correct answer, it was added to the list.

\subsection{Postprocessing}

\begin{table*}[!htbp]
\centering
\small
\begin{tabular}{@{}ccccc@{}}
\toprule
\textbf{Dataset} & \textbf{\begin{tabular}[c]{@{}c@{}}Total\\ (train / val / test)\end{tabular}} & \textbf{\begin{tabular}[c]{@{}c@{}}Single-Token\\ (train / val / test)\end{tabular}} & \textbf{\begin{tabular}[c]{@{}c@{}}Answer in Short\\ (train / val / test)\end{tabular}} & \textbf{\begin{tabular}[c]{@{}c@{}}Answer in Long\\ (train / val / test)\end{tabular}} \\ \midrule
\textsc{Quasar-S}         & 31,049 / 3,174 / 3,139                                                        & --                                                                                   & 30,198 / 3,084 / 3,044                                                                  & 30,417 / 3,099 / 3,064                                                                 \\
\textsc{Quasar-T}         & 37,012 / 3,000 / 3,000                                                        & 18,726 / 1,507 / 1,508                                                               & 25,465 / 2,068 / 2,043                                                                  & 26,318 / 2,129 / 2,102                                                                 \\ \bottomrule
\end{tabular}
\caption{\small Dataset Statistics. \textbf{Single-Token} refers to the questions whose answer is a single token (for \textsc{Quasar-S} all answers come from a fixed vocabulary). \textbf{Answer in Short (Long)} indicates whether the answer is present in the retrieved short (long) pseudo-documents.}
\label{t_datasetStats}
\end{table*}

% \begin{table*}[!htbp]
% \centering
% \small
% \begin{tabular}{|l|c|c|c|c|}
% \hline
% \textbf{Dataset} & \textbf{Pseudodoc size} & \textbf{Total entries} & \textbf{Answer included} & \textbf{Train / Val / Test}\\\hline
% Software & Large & 37,362 & 36,580 & 31,049 / 3,139 / 3,174\\\hline
% Software & Small & 37,362 & 36,326 & 31,049 / 3,139 / 3,174\\\hline
% % Trivia (all)  & Large & 45,060 & 33,326 &\\\hline
% Trivia (all)  & Large & 43,012 & 33,326 & 37,012 / 3,000 / 3,000\\\hline
% Trivia (all)  & Small & 43,012 & 29,598 & 37,012 / 3,000 / 3,000\\\hline
% % Trivia (1tok) & Large & 23,163 & 20,403 &\\\hline
% Trivia (1tok) & Large & 21,741 & 18,712 & 18,726 / 1,508 / 1,507\\\hline
% Trivia (1tok) & Small & 21,741 & 18,377 & 18,726 / 1,508 / 1,507\\\hline
% \end{tabular}
% \caption{\small Dataset statistics}\label{t_datasetStats}
% \end{table*}

Once context documents had been built, we extracted the subset of questions where the answer string, excluded from the query for the two-phase search, was nonetheless present in the context document. This subset allows us to evaluate the performance of the reading system independently from the search system, while the full set allows us to evaluate the performance of \textsc{Quasar} as a whole. We also split the full set into training, validation and test sets. The final size of each data subset after all discards is listed in Table \ref{t_datasetStats}.

\section{Evaluation}
\subsection{Metrics}
Evaluation is straightforward on \textsc{Quasar-S} since each answer comes from a fixed output vocabulary of entities, and we report the average \textit{accuracy} of predictions as the evaluation metric. For \textsc{Quasar-T}, the answers may be free form spans of text, and the same answer may be expressed in different terms, which makes evaluation difficult. Here we pick the two metrics from \citet{rajpurkar2016squad,joshi2017triviaqa}.
In preprocessing the answer we remove punctuation, white-space and definite and indefinite articles from the strings. Then, \textit{exact} match measures whether the two strings, after preprocessing, are equal or not. For \textit{F1} match we first construct a bag of tokens for each string, followed be preprocessing of each token, and measure the F1 score of the overlap between the two bags of tokens. These metrics are far from perfect for \textsc{Quasar-T}; for example, our human testers were penalized for entering ``0'' as answer instead of ``zero''. However, a comparison between systems may still be meaningful.

\subsection{Human Evaluation}
To put the difficulty of the introduced datasets into perspective, we evaluated human performance on answering the questions. For each dataset, we recruited one domain expert (a developer with several years of programming experience for \textsc{Quasar-S}, and an avid trivia enthusiast for \textsc{Quasar-T}) and $1-3$  non-experts. Each volunteer was presented with randomly selected questions from the development set and asked to answer them via an online app. The experts were evaluated in a ``closed-book'' setting, i.e. they did not have access to any external resources. The non-experts were evaluated in an ``open-book'' setting, where they had access to a search engine over the short pseudo-documents extracted for each dataset (as described in Section~\ref{sec:retreival}). We decided to use short pseudo-documents for this exercise to reduce the burden of reading on the volunteers, though we note that the long pseudo-documents have greater coverage of answers. 

\begin{figure*}
\centering
    \begin{subfigure}[b]{0.32\textwidth}
        \includegraphics[width=\textwidth,trim={5mm 0 5mm 0},clip]{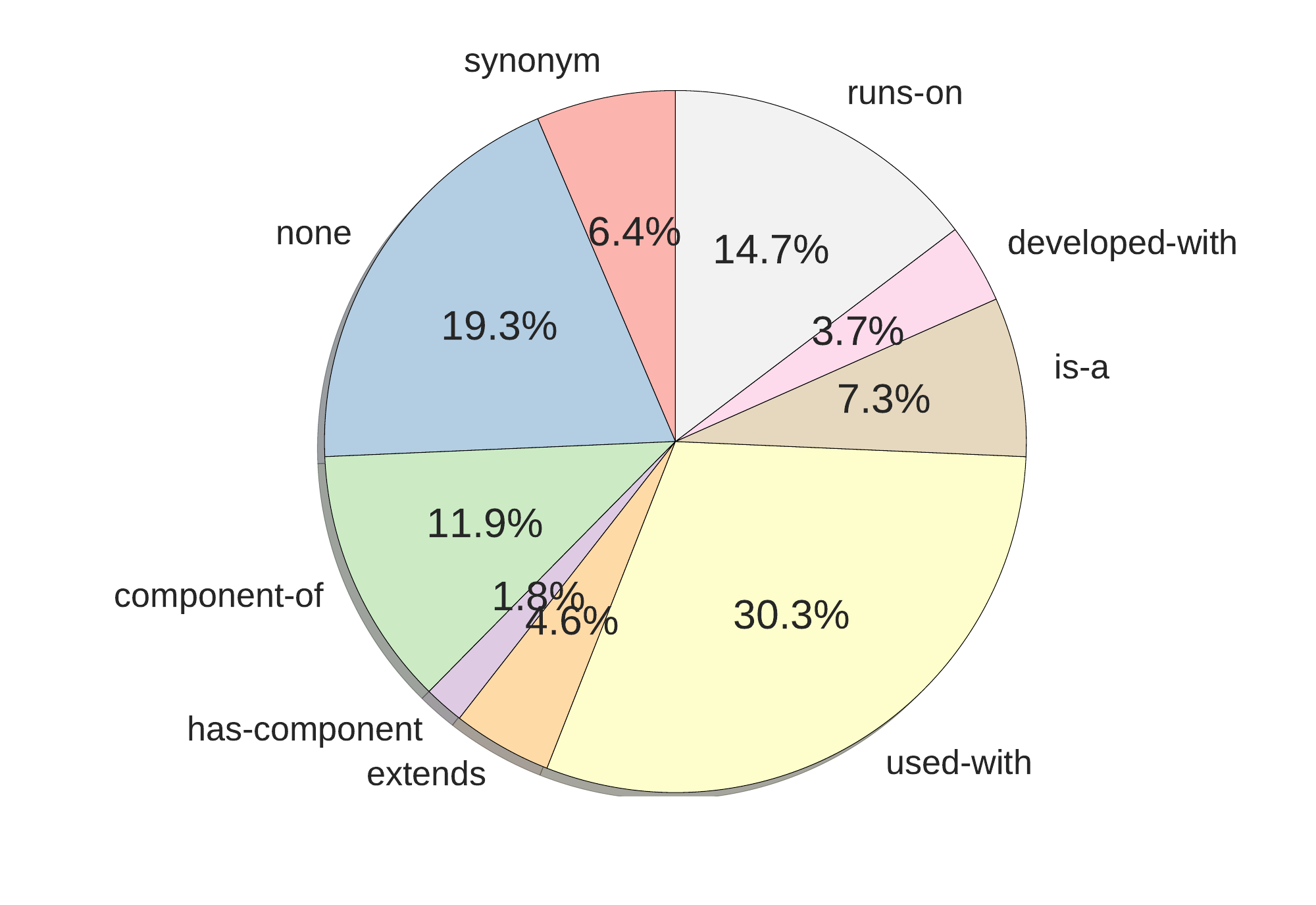}
        \caption{\textsc{Quasar-S} relations}
        \label{fig:so_rel}
    \end{subfigure}
    \begin{subfigure}[b]{0.32\textwidth}
        \includegraphics[width=\textwidth,trim={5mm 0 5mm 0},clip]{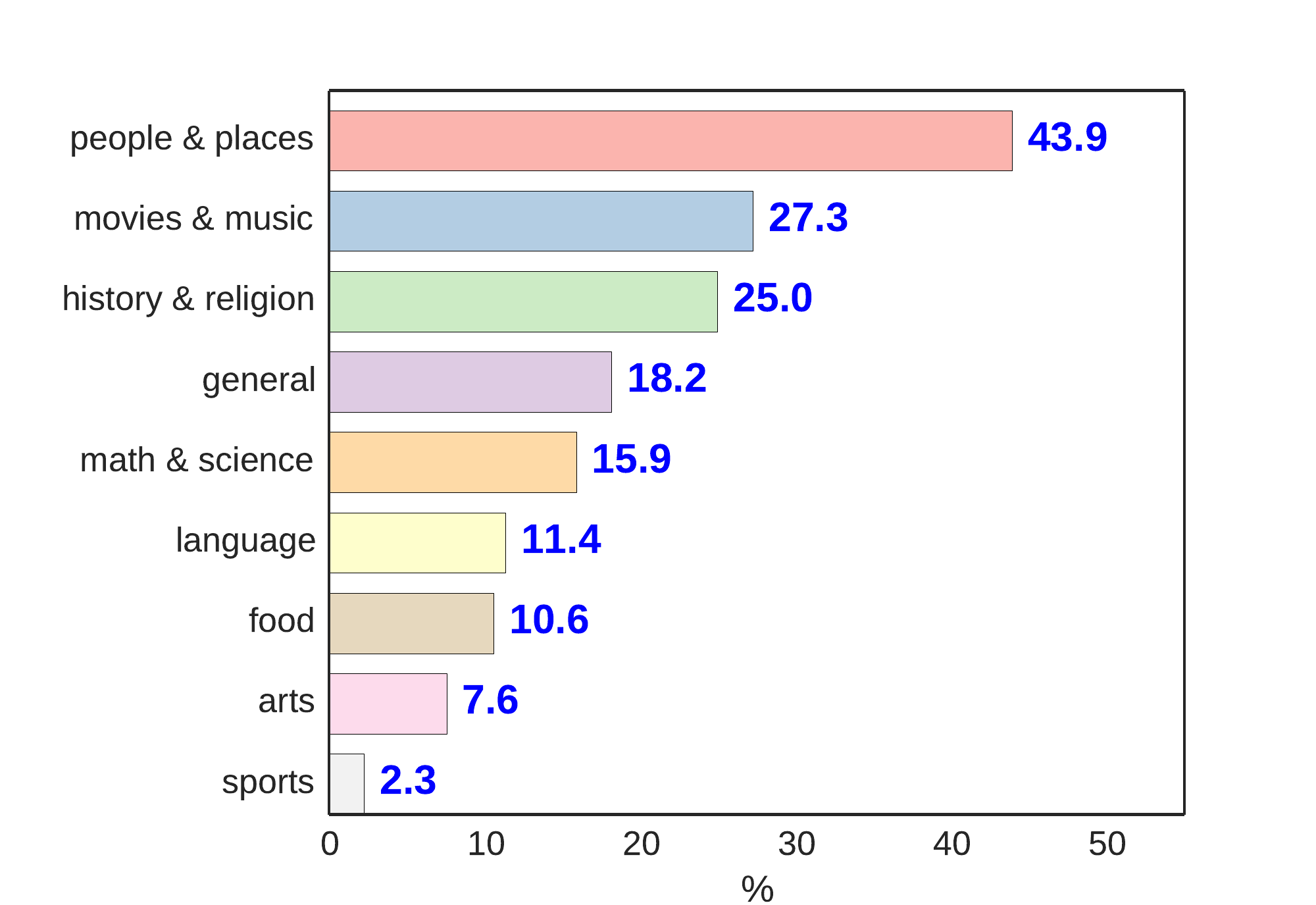}
        \caption{\textsc{Quasar-T} genres}
        \label{fig:tr_rel}
    \end{subfigure}
    \begin{subfigure}[b]{0.32\textwidth}
        \includegraphics[width=\textwidth,trim={5mm 0 5mm 0},clip]{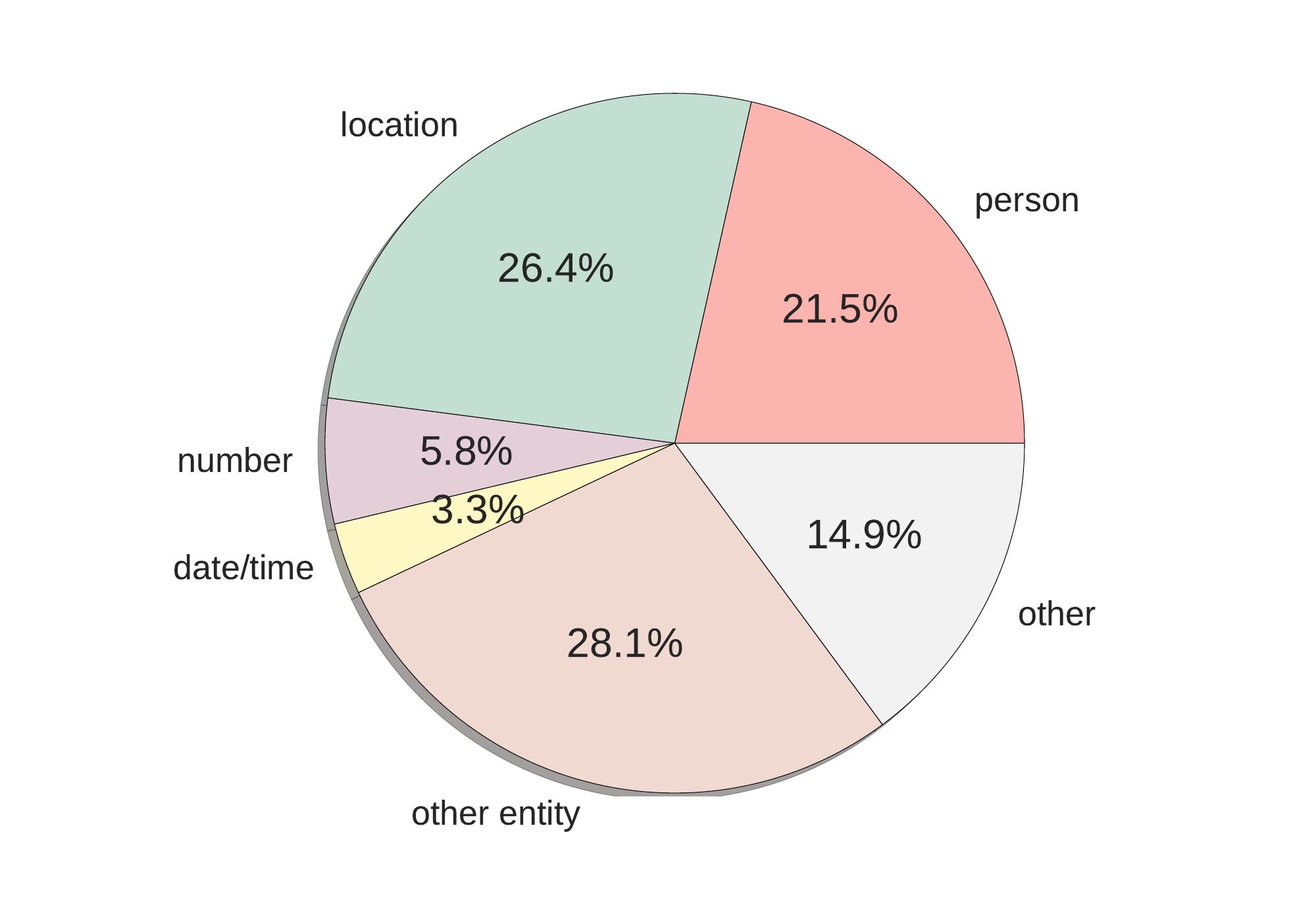}
        \caption{\textsc{Quasar-T} answer categories}
        \label{fig:tr_typ}
    \end{subfigure}
    \caption{\small Distribution of manual annotations for \textsc{Quasar}. Description of the \textsc{Quasar-S} annotations is in Appendix~\ref{app:relations}.}\label{fig:annotations}
\end{figure*}

We also asked the volunteers to provide annotations to categorize the type of each question they were asked, and a label for whether the question was ambiguous. For \textsc{Quasar-S} the annotators were asked to mark the relation between the \textit{head} entity (from whose definition the cloze was constructed) and the \textit{answer} entity. For \textsc{Quasar-T} the annotators were asked to mark the genre of the question (e.g., Arts \& Literature)\footnote{Multiple genres per question were allowed.} and the entity type of the answer (e.g., Person). When multiple annotators marked the same question differently, we took the majority vote when possible and discarded ties. In total we collected $226$ relation annotations for $136$ questions in \textsc{Quasar-S}, out of which $27$ were discarded due to conflicting ties, leaving a total of $109$ annotated questions. For \textsc{Quasar-T} we collected annotations for a total of $144$ questions, out of which $12$ we marked as ambiguous. In the remaining $132$, a total of $214$ genres were annotated (a question could be annotated with multiple genres), while $10$ questions had conflicting entity-type annotations which we discarded, leaving $122$ total entity-type annotations. Figure~\ref{fig:annotations} shows the distribution of these annotations. 

\subsection{Baseline Systems}
We evaluate several baselines on \textsc{Quasar}, ranging from simple heuristics to deep neural networks. Some predict a single token / entity as the answer, while others predict a span of tokens. 
% The baselines fall into two categories -- \textit{comprehension} baselines which extract the answer from the retrieved context, and \textit{language-modeling} baselines (only for \textsc{Quasar-S}) which are trained on the entire corpus and directly answer the cloze-style questions.

\subsubsection{Heuristic Models}
\paragraph{Single-Token:} \textit{MF-i} (Maximum Frequency) counts the number of occurrences of each candidate answer in the retrieved context and returns the one with maximum frequency. \textit{MF-e} is the same as \textit{MF-i} except it excludes the candidates present in the query. \textit{WD} (Word Distance) measures the sum of distances from a candidate to other non-stopword tokens in the passage which are also present in the query. For the cloze-style \textsc{Quasar-S} the distances are measured by first aligning the query placeholder to the candidate in the passage, and then measuring the offsets between other tokens in the query and their mentions in the passage. The maximum distance for any token is capped at a specified threshold, which is tuned on the validation set. 

\paragraph{Multi-Token:} For \textsc{Quasar-T} we also test the Sliding Window (\textit{SW}) and Sliding Window + Distance (\textit{SW+D}) baselines proposed in \citep{richardson2013mctest}. The scores were computed for the list of candidate solutions described in Section~\ref{sec:retreival}.

\subsubsection{Language Models}
For \textsc{Quasar-S}, since the answers come from a fixed vocabulary of entities, we test language model baselines which predict the most likely entity to appear in a given context. We train three \textit{n-gram} baselines using the SRILM toolkit \citep{stolcke2002srilm} for $n=3,4,5$ on the entire corpus of all Stack Overflow posts. The output predictions are restricted to the output vocabulary of entities.

We also train a bidirectional Recurrent Neural Network (RNN) language model (based on GRU units). This model encodes both the left and right context of an entity using forward and backward GRUs, and then concatenates the final states from both to predict the entity through a softmax layer. Training is performed on the entire corpus of Stack Overflow posts, with the loss computed only over mentions of entities in the output vocabulary. This approach benefits from looking at both sides of the cloze in a query to predict the entity, as compared to the single-sided n-gram baselines.

\subsubsection{Reading Comprehension Models}
Reading comprehension models are trained to extract the answer from the given passage. We test two recent architectures on \textsc{Quasar} using publicly available code from the authors\footnote{\url{https://github.com/bdhingra/ga-reader}} \footnote{\url{https://github.com/allenai/bi-att-flow}}.

\paragraph{GA (Single-Token):} The GA Reader \citep{dhingra2016gated} is a multi-layer neural network which extracts a single token from the passage to answer a given query. At the time of writing it had state-of-the-art performance on several cloze-style datasets for QA. For \textsc{Quasar-S} we train and test GA on all instances for which the correct answer is found within the retrieved context. For \textsc{Quasar-T} we train and test GA on all instances where the answer is in the context and is a single token.

\paragraph{BiDAF (Multi-Token):} The BiDAF model \citep{seo2016bidirectional} is also a multi-layer neural network which predicts a span of text from the passage as the answer to a given query. At the time of writing it had state-of-the-art performance among published models on the Squad dataset. For \textsc{Quasar-T} we train and test BiDAF on all instances where the answer is in the retrieved context. 

\subsection{Results}
\begin{figure*}
\centering
    \begin{subfigure}[b]{0.24\textwidth}
        \includegraphics[width=\textwidth,trim={5mm 0 5mm 0},clip]{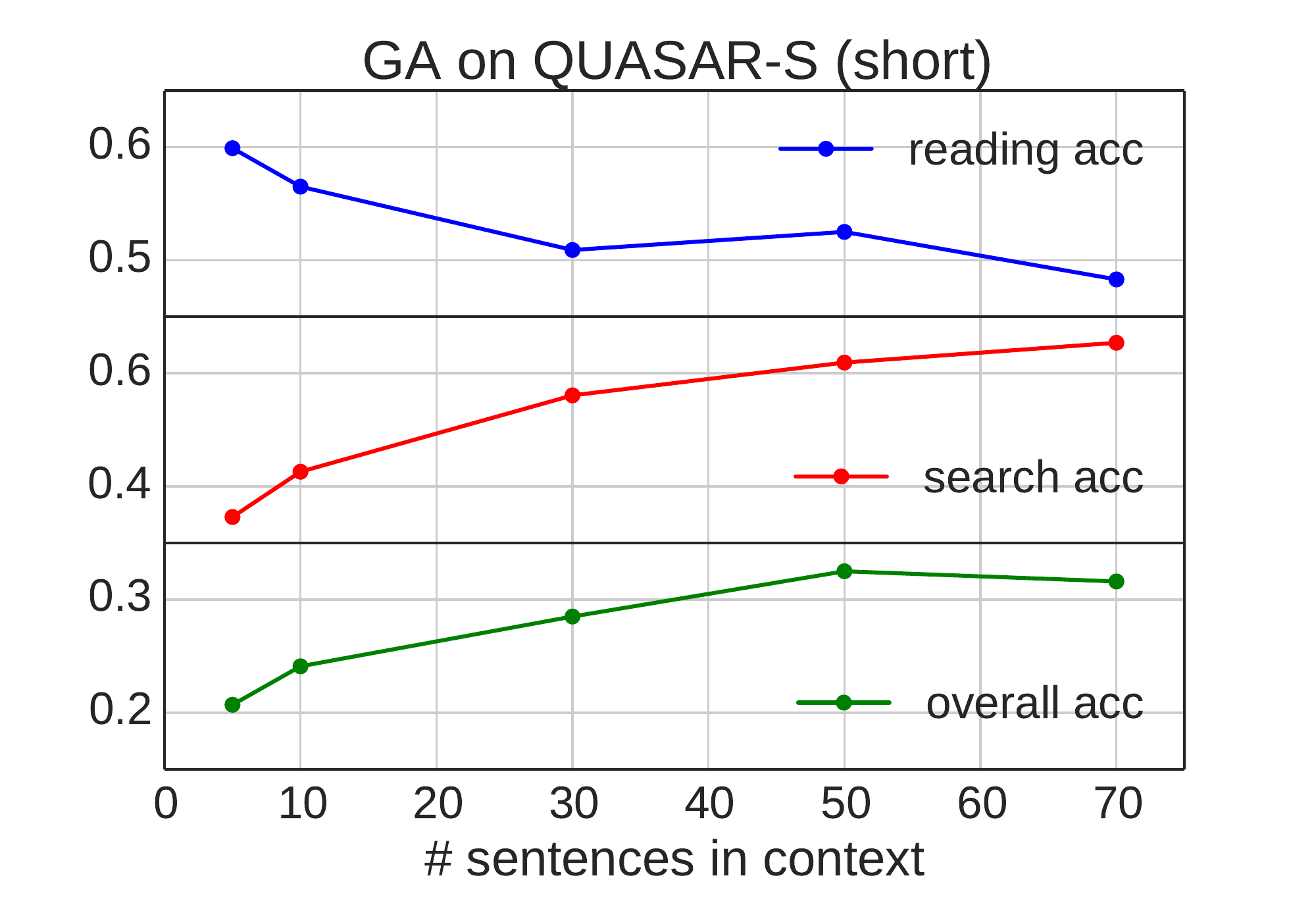}
    \end{subfigure}
    \begin{subfigure}[b]{0.24\textwidth}
        \includegraphics[width=\textwidth,trim={5mm 0 5mm 0},clip]{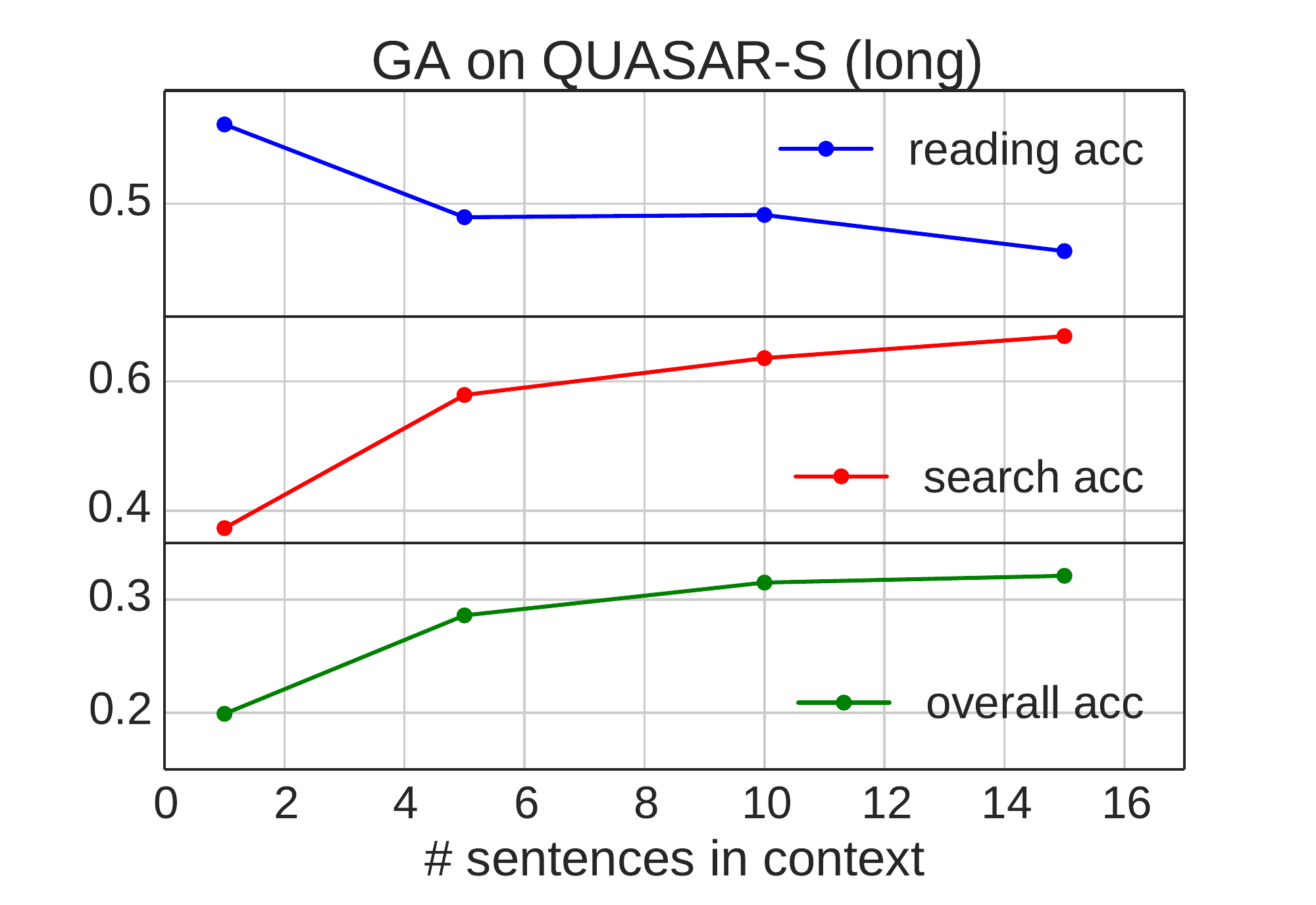}
    \end{subfigure}
    \begin{subfigure}[b]{0.24\textwidth}
        \includegraphics[width=\textwidth,trim={5mm 0 5mm 0},clip]{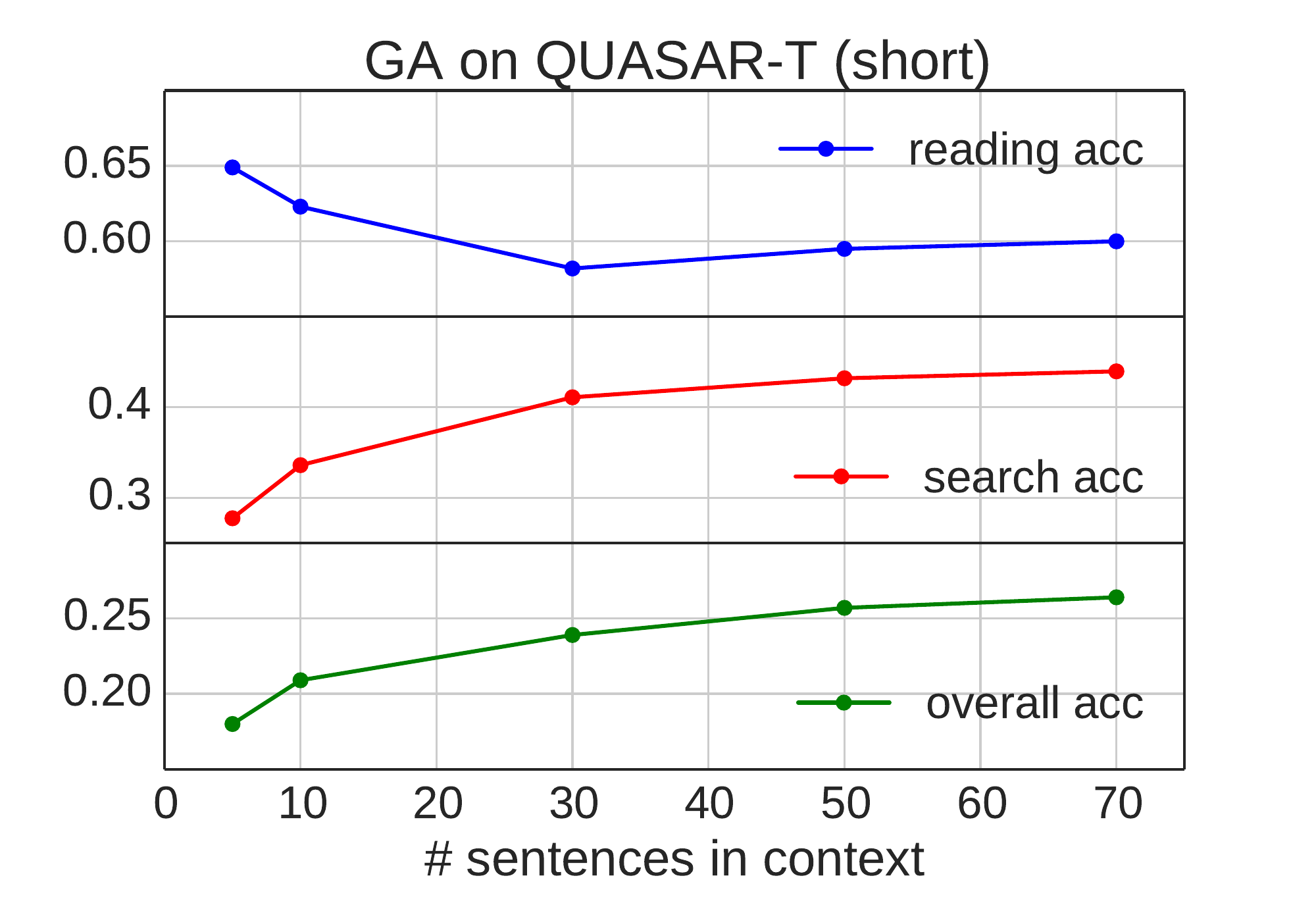}
    \end{subfigure}
    \begin{subfigure}[b]{0.24\textwidth}
        \includegraphics[width=\textwidth,trim={5mm 0 5mm 0},clip]{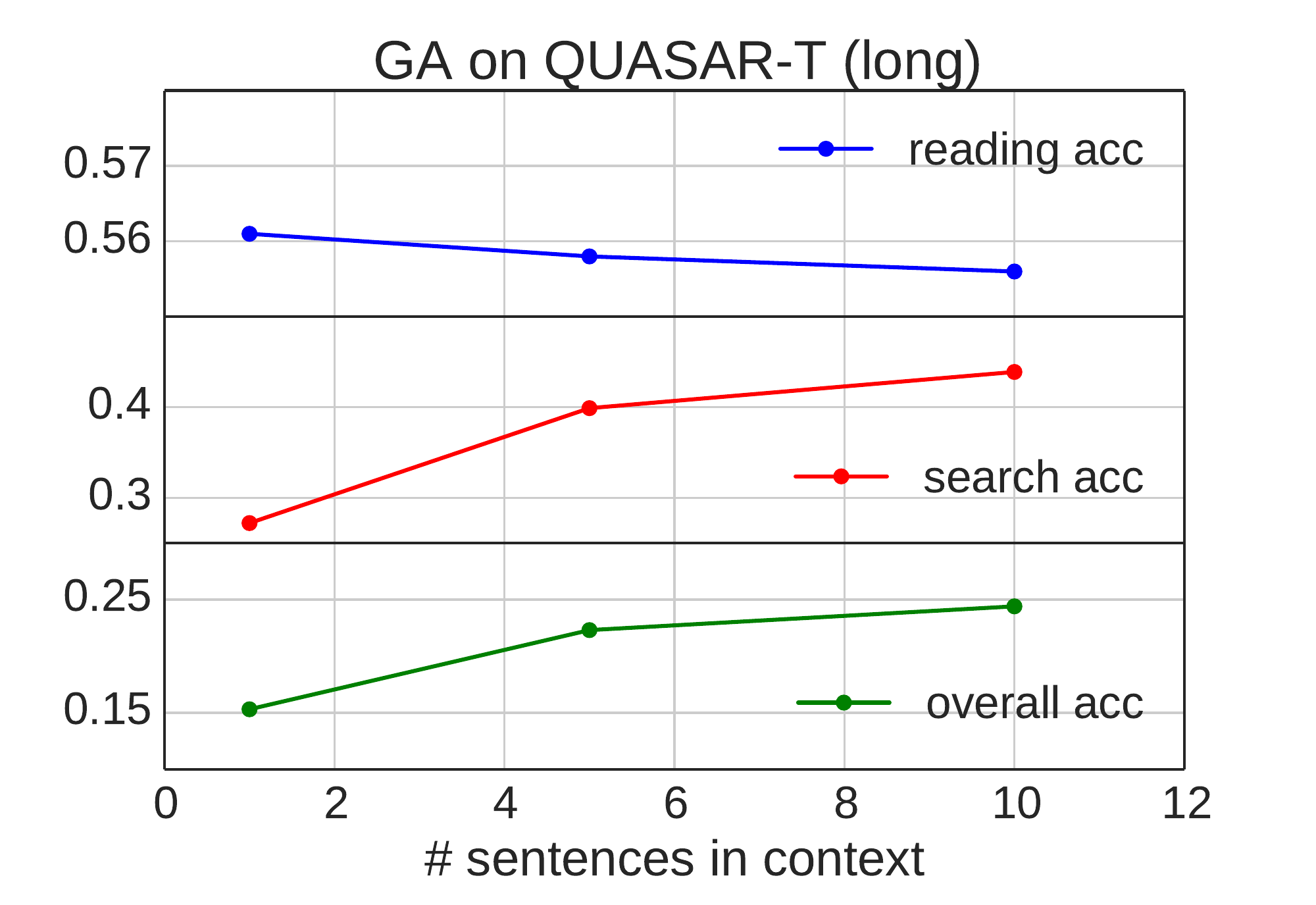}
    \end{subfigure}
    \caption{\small Variation of Search, Read and Overall accuracies as the number of context documents is varied.}\label{fig:searchvread}
\end{figure*}

Several baselines rely on the retrieved context to extract the answer to a question. For these, we refer to the fraction of instances for which the correct answer is present in the context as \textit{Search Accuracy}. The performance of the baseline among these instances is referred to as the \textit{Reading Accuracy}, and the overall performance (which is a product of the two) is referred to as the \textit{Overall Accuracy}. In Figure~\ref{fig:searchvread} we compare how these three vary as the number of context documents is varied. Naturally, the search accuracy increases as the context size increases, however at the same time reading performance decreases since the task of extracting the answer becomes harder for longer documents. Hence, simply retrieving more documents is not sufficient -- finding the few most relevant ones will allow the reader to work best.

\begin{table*}[!htbp]
\small
\centering
\begin{tabular}{|l|c|c|c|c|c|c|c|}
\hline
\multirow{2}{*}{\textbf{Method}} & \multirow{2}{*}{\textbf{\begin{tabular}[c]{@{}c@{}}Optimal\\ Context\end{tabular}}} & \multicolumn{2}{c|}{\textbf{Search Acc}} & \multicolumn{2}{c|}{\textbf{Reading Acc}} & \multicolumn{2}{c|}{\textbf{Overall Acc}} \\ \cline{3-8} 
                                 &                                                                                     & \textbf{val}          & \textbf{test}         & \textbf{val}           & \textbf{test}         & \textbf{val}           & \textbf{test}         \\ \hline
\multicolumn{8}{|l|}{Human Performance}                                                                                                                                                                                                                                  \\ \hline
Expert (CB)                      & --                                                                                  & --                    & --                    & --                     & --                    & \textit{0.468}                     & --        \\
Non-Expert (OB)                  & --                                                                                  & --                    & --                    & --                     & --                    & \textit{0.500}                     & --        \\ \hline
\multicolumn{8}{|l|}{Language models}                                                                                                                                                                                                                                    \\ \hline
3-gram                           & --                                                                                  & --                    & --                    & --                     & --                    & 0.148                  & 0.153                 \\
4-gram                           & --                                                                                  & --                    & --                    & --                     & --                    & 0.161                  & 0.171                 \\
5-gram                           & --                                                                                  & --                    & --                    & --                     & --                    & 0.165                  & 0.174                 \\
BiRNN$\dagger$                            & --                                                                                  & --                    & --                    & --                     & --                    & \textbf{0.345}         & \textbf{0.336}        \\ \hline
\multicolumn{8}{|l|}{Short-documents}                                                                                                                                                                                                                                    \\ \hline
WD                               & 10                                                                                  & 0.40                  & 0.43                  & 0.250                  & 0.249                 & 0.100                  & 0.107                 \\
MF-e                             & 60                                                                                  & 0.64                  & 0.64                  & 0.209                  & 0.212                 & 0.134                  & 0.136                 \\
MF-i                             & 90                                                                                  & 0.67                  & 0.68                  & 0.237                  & 0.234                 & 0.159                  & 0.159                 \\
GA$\dagger$                               & 70                                                                                  & 0.65                  & 0.65                  & \textbf{0.486}         & \textbf{0.483}        & 0.315                  & 0.316                 \\ \hline
\multicolumn{8}{|l|}{Long-documents}                                                                                                                                                                                                                                     \\ \hline
WD                               & 10                                                                                  & 0.66                  & 0.66                  & 0.124                  & 0.142                 & 0.082                  & 0.093                 \\
MF-e                             & 15                                                                                  & 0.69                  & 0.69                  & 0.185                  & 0.197                 & 0.128                  & 0.136                 \\
MF-i                             & 15                                                                                  & 0.69                  & 0.69                  & 0.230                  & 0.231                 & 0.159                  & 0.159                 \\
GA$\dagger$                               & 15                                                                                  & 0.67                  & 0.67                  & 0.474                  & 0.479                 & 0.318                  & 0.321                 \\ \hline
\end{tabular}
\caption{\small Performance comparison on \textsc{Quasar-S}. CB: Closed-Book, OB: Open Book. Neural baselines are denoted with $\dagger$. Optimal context is the number of documents used for answer extraction, which was tuned to maximize the overall accuracy on validation set.}
\label{tab:quasars}
\end{table*}

\begin{table*}[!htbp]
\small
\centering
\begin{tabular}{|l|c|c|c|c|c|c|c|c|c|c|c|}
\hline
\multirow{3}{*}{\textbf{Method}} & \multirow{3}{*}{\textbf{\begin{tabular}[c]{@{}c@{}}Optimal \\ Context\end{tabular}}} & \multicolumn{2}{c|}{\multirow{2}{*}{\textbf{\begin{tabular}[c]{@{}c@{}}Search\\ Acc\end{tabular}}}} & \multicolumn{4}{c|}{\textbf{Reading Acc}}                              & \multicolumn{4}{c|}{\textbf{Overall Acc}}                              \\ \cline{5-12} 
                                 &                                                                                      & \multicolumn{2}{c|}{}                                                                               & \multicolumn{2}{c|}{\textbf{exact}} & \multicolumn{2}{c|}{\textbf{f1}} & \multicolumn{2}{c|}{\textbf{exact}} & \multicolumn{2}{c|}{\textbf{f1}} \\ \cline{3-12} 
                                 &                                                                                      & \textbf{val}                                     & \textbf{test}                                    & \textbf{val}     & \textbf{test}    & \textbf{val}    & \textbf{test}  & \textbf{val}     & \textbf{test}    & \textbf{val}    & \textbf{test}  \\ \hline
\multicolumn{12}{|l|}{Human Performance}                                                                                                                                                                                                                                                                                                                                        \\ \hline
Expert (CB)                      & --                                                                                   & --                                               & --                                               & --               & --               & --              & --             & \textit{0.547}               & --   & \textit{0.604}              & -- \\
Non-Expert (OB)                  & --                                                                                   & --                                               & --                                               & --               & --               & --              & --             & \textit{0.515}               & --   & \textit{0.606}              & -- \\ \hline
\multicolumn{12}{|l|}{Short-documents}                                                                                                                                                                                                                                                                                                                                          \\ \hline
MF-i                             & 10                                                                                   & 0.35                                             & 0.34                                             & 0.053            & 0.044            & 0.053           & 0.044          & 0.019            & 0.015            & 0.019           & 0.015          \\
WD                               & 20                                                                                   & 0.40                                             & 0.39                                             & 0.104            & 0.082            & 0.104           & 0.082          & 0.042            & 0.032            & 0.042           & 0.032          \\
SW+D                             & 20                                                                                   & 0.64                                             & 0.63                                             & 0.112            & 0.113            & 0.157           & 0.155          & 0.072            & 0.071            & 0.101           & 0.097          \\
SW                               & 10                                                                                   & 0.56                                             & 0.53                                             & 0.216            & 0.205            & 0.299           & 0.271          & 0.120            & 0.109            & 0.159           & 0.144          \\
MF-e                             & 70                                                                                   & 0.45                                             & 0.45                                             & 0.372            & 0.342            & 0.372           & 0.342          & 0.167            & 0.153            & 0.167           & 0.153          \\
GA$\dagger$                               & 70                                                                                   & 0.44                                             & 0.44                                             & \textbf{0.580}   & \textbf{0.600}   & \textbf{0.580}  & \textbf{0.600} & 0.256            & \textbf{0.264}   & 0.256           & 0.264          \\
BiDAF$\dagger$**                            & 10                                                                                   & 0.57                                             & 0.54                                             & 0.454            & 0.476            & 0.509           & 0.524          & \textbf{0.257}   & 0.259            & \textbf{0.289}  & \textbf{0.285} \\ \hline
\multicolumn{12}{|l|}{Long-documents}                                                                                                                                                                                                                                                                                                                                           \\ \hline
WD                               & 20                                                                                   & 0.43                                             & 0.44                                             & 0.084            & 0.067            & 0.084           & 0.067          & 0.037            & 0.029            & 0.037           & 0.029          \\
SW                               & 20                                                                                   & 0.74                                             & 0.73                                             & 0.041            & 0.034            & 0.056           & 0.050          & 0.030            & 0.025            & 0.041           & 0.037          \\
SW+D                             & 5                                                                                    & 0.58                                             & 0.58                                             & 0.064            & 0.055            & 0.094           & 0.088          & 0.037            & 0.032            & 0.054           & 0.051          \\
MF-i                             & 20                                                                                   & 0.44                                             & 0.45                                             & 0.185            & 0.187            & 0.185           & 0.187          & 0.082            & 0.084            & 0.082           & 0.084          \\
MF-e                             & 20                                                                                   & 0.43                                             & 0.44                                             & 0.273            & 0.286            & 0.273           & 0.286          & 0.119            & 0.126            & 0.119           & 0.126          \\
BiDAF$\dagger$**                            & 1                                                                                    & 0.47                                            & 0.468                                            & 0.370            & 0.395            & 0.425           & 0.445          & 0.17            & 0.185            & 0.199           & 0.208          \\
GA$\dagger$**                               & 10                                                                                   & 0.44                                            & 0.44                                            & 0.551            & 0.556            & 0.551           & 0.556          & 0.245            & 0.244            & 0.245           & 0.244          \\ \hline
\end{tabular}
\caption{\small Performance comparison on \textsc{Quasar-T}. CB: Closed-Book, OB: Open Book. Neural baselines are denoted with $\dagger$. Optimal context is the number of documents used for answer extraction, which was tuned to maximize the overall accuracy on validation set.**We were unable to run BiDAF with more than 10 short-documents / 1 long-documents, and GA with more than 10 long-documents due to memory errors.}
\label{tab:quasart}
\end{table*}

In Tables~\ref{tab:quasars} and \ref{tab:quasart} we compare all baselines when the context size is tuned to maximize the overall accuracy on the validation set\footnote{The Search Accuracy for different baselines may be different despite the same number of retrieved context documents, due to different preprocessing requirements. For example, the SW baselines allow multiple tokens as answer, whereas WD and MF baselines do not.}. For \textsc{Quasar-S} the best performing baseline is the BiRNN language model, which achieves $33.6\%$ accuracy. The GA model achieves $48.3\%$ accuracy on the set of instances for which the answer is in context, however, a search accuracy of only $65\%$ means its overall performance is lower. This can improve with improved retrieval. For \textsc{Quasar-T}, both the neural models significantly outperform the heuristic models, with BiDAF getting the highest F1 score of $28.5\%$. 

The best performing baselines, however, lag behind human performance by $16.4\%$ and $32.1\%$ for \textsc{Quasar-S} and \textsc{Quasar-T} respectively, indicating the strong potential for improvement. Interestingly, for human performance we observe that non-experts are able to match or beat the performance of experts when given access to the background corpus for searching the answers. We also emphasize that the human performance is limited by either the knowledge of the experts, or the usefulness of the search engine for non-experts; it should not be viewed as an upper bound for automatic systems which can potentially use the entire background corpus. 
Further analysis of the human and baseline performance in each category of annotated questions is provided in Appendix~\ref{app:performance}.

\section{Conclusion}
We have presented the \textsc{Quasar} datasets for promoting research into two related tasks for QA -- searching a large corpus of text for relevant passages, and reading the passages to extract answers. We have also described baseline systems for the two tasks which perform reasonably but lag behind human performance. While the searching performance improves as we retrieve more context, the reading performance typically goes down. Hence, future work, in addition to improving these components individually, should also focus on joint approaches to optimizing the two on end-task performance. The datasets, including the documents retrieved by our system and the human annotations, are available at \url{https://github.com/bdhingra/quasar}. 

\section*{Acknowledgments}
This work was funded by NSF under grants CCF-1414030 and IIS-1250956 and by grants from Google.

\bibliographystyle{emnlp_natbib}
\bibliography{emnlp2017}

\appendix

\section{\textsc{Quasar-S} Relation Definitions}
\label{app:relations}

\begin{table*}[!htbp]
\centering
\small
\begin{tabular}{@{}l|l@{}}
\toprule
\textbf{\begin{tabular}[c]{@{}l@{}}Relation \\ (head $\to$ answer)\end{tabular}} & \textbf{Definition}                                                                                                     \\ \midrule
is-a                                                                             & \textit{head} is a type of \textit{answer}                                                                                                \\
component-of                                                                     & \textit{head} is a component of \textit{answer}                                                                                           \\
has-component                                                                    & \textit{answer} is a component of \textit{head}                                                                                           \\
developed-with                                                                   & \textit{head} was developed using the \textit{answer}                                                                                     \\
extends                                                                          & \textit{head} is a plugin or library providing additional functionality to larger thing \textit{answer}        \\
runs-on                                                                          & \textit{answer} is an operating system, platform, or framework on which \textit{head} runs         \\
synonym                                                                          & \textit{head} and \textit{answer} are the same entity                                                                                      \\
used-for                                                                         & \textit{head} is a software / framework used for some functionality related to \textit{answer} \\ \bottomrule
\end{tabular}
\caption{\small Description of the annotated relations between the \textit{head} entity, from whose definition the cloze is constructed, and the \textit{answer} entity which fills in the cloze. These are the same as the descriptions shown to the annotators. }
\label{tab:descriptions}
\end{table*}

\begin{figure*}[!htbp]
\centering
    \begin{subfigure}[b]{0.32\textwidth}
        \includegraphics[width=\textwidth,trim={5mm 0 5mm 0},clip]{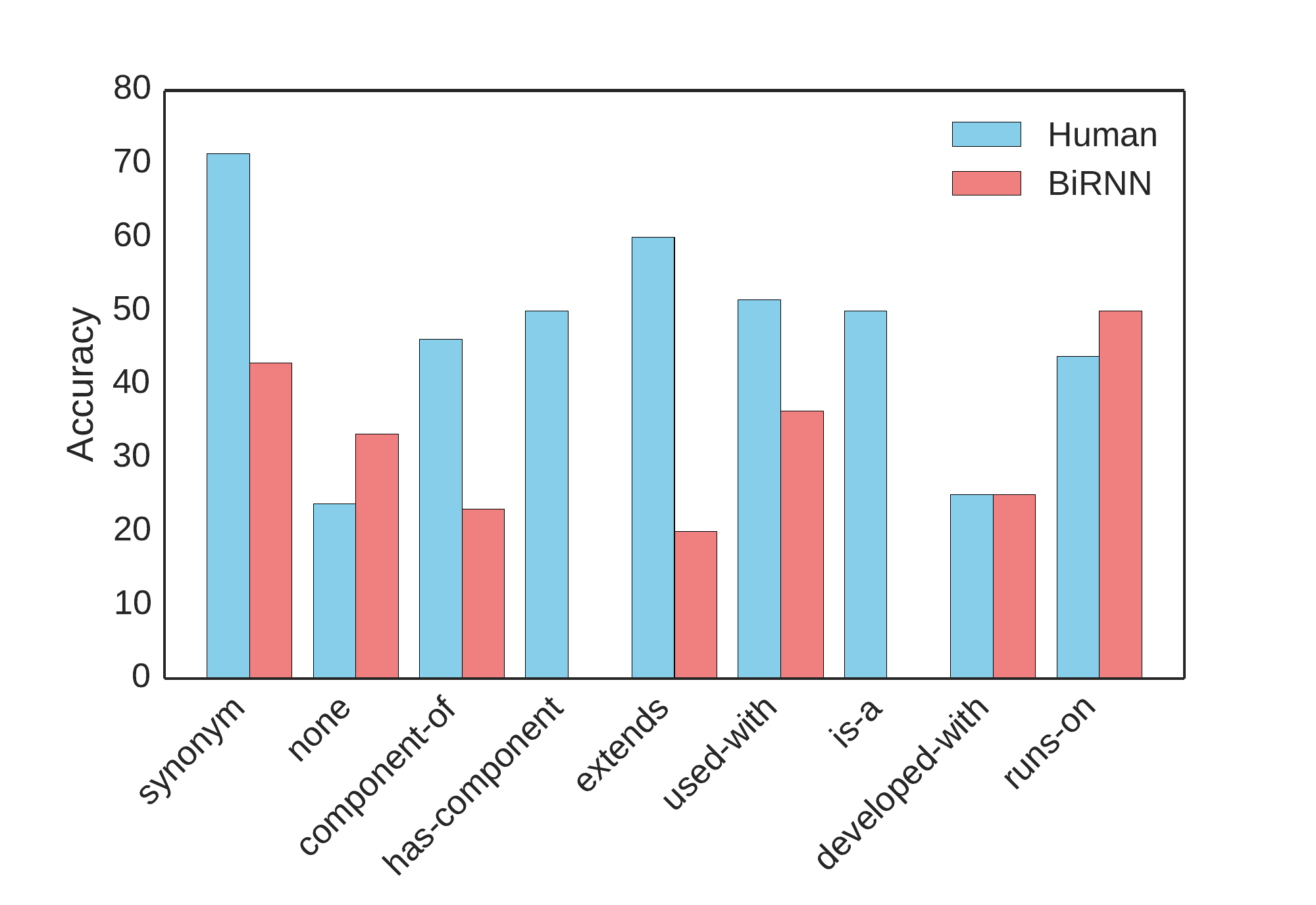}
        \caption{\textsc{Quasar-S} relations}
        \label{fig:so_rel}
    \end{subfigure}
    \begin{subfigure}[b]{0.32\textwidth}
        \includegraphics[width=\textwidth,trim={5mm 0 5mm 0},clip]{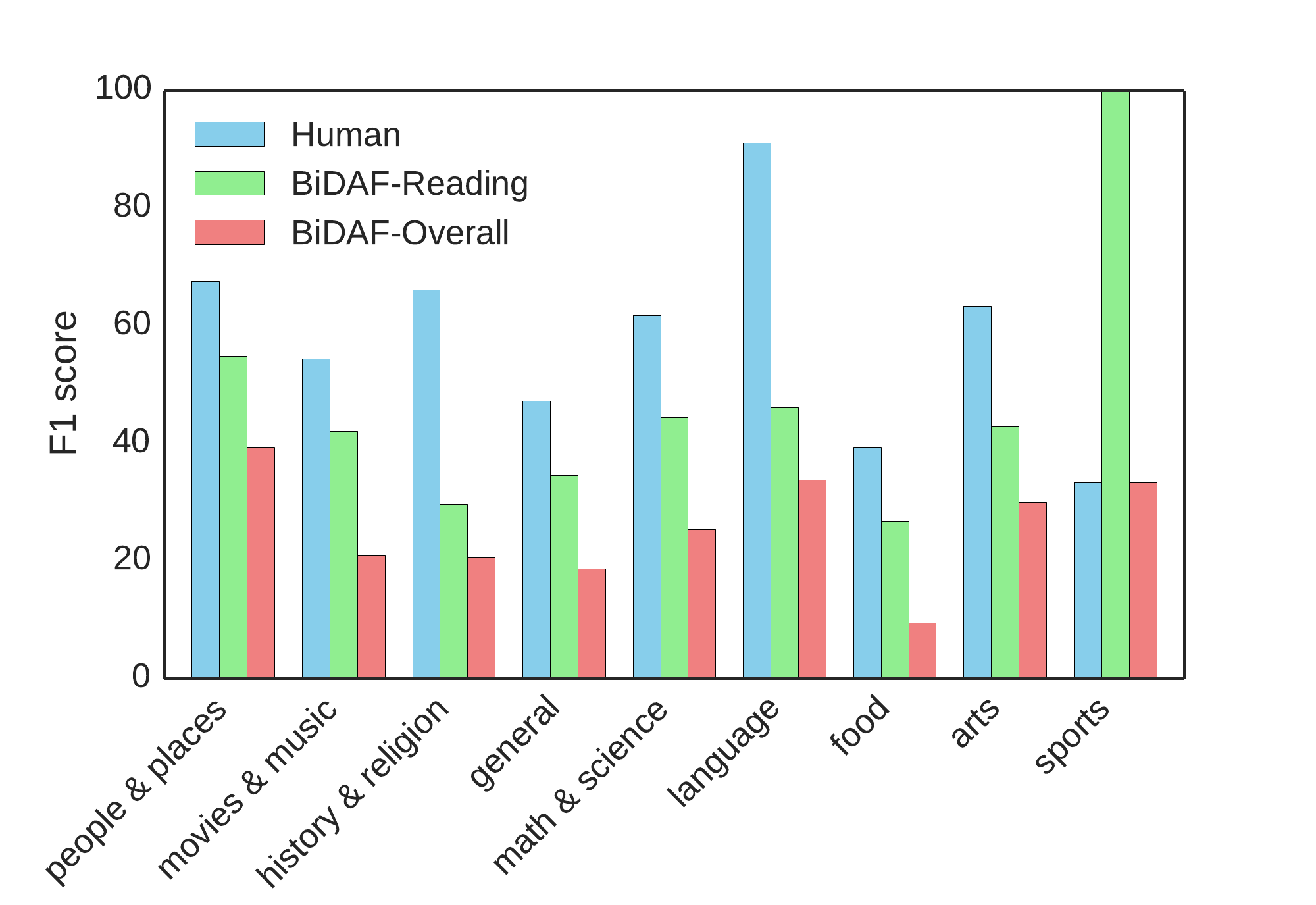}
        \caption{\textsc{Quasar-T} genres}
        \label{fig:tr_rel}
    \end{subfigure}
    \begin{subfigure}[b]{0.32\textwidth}
        \includegraphics[width=\textwidth,trim={5mm 0 5mm 0},clip]{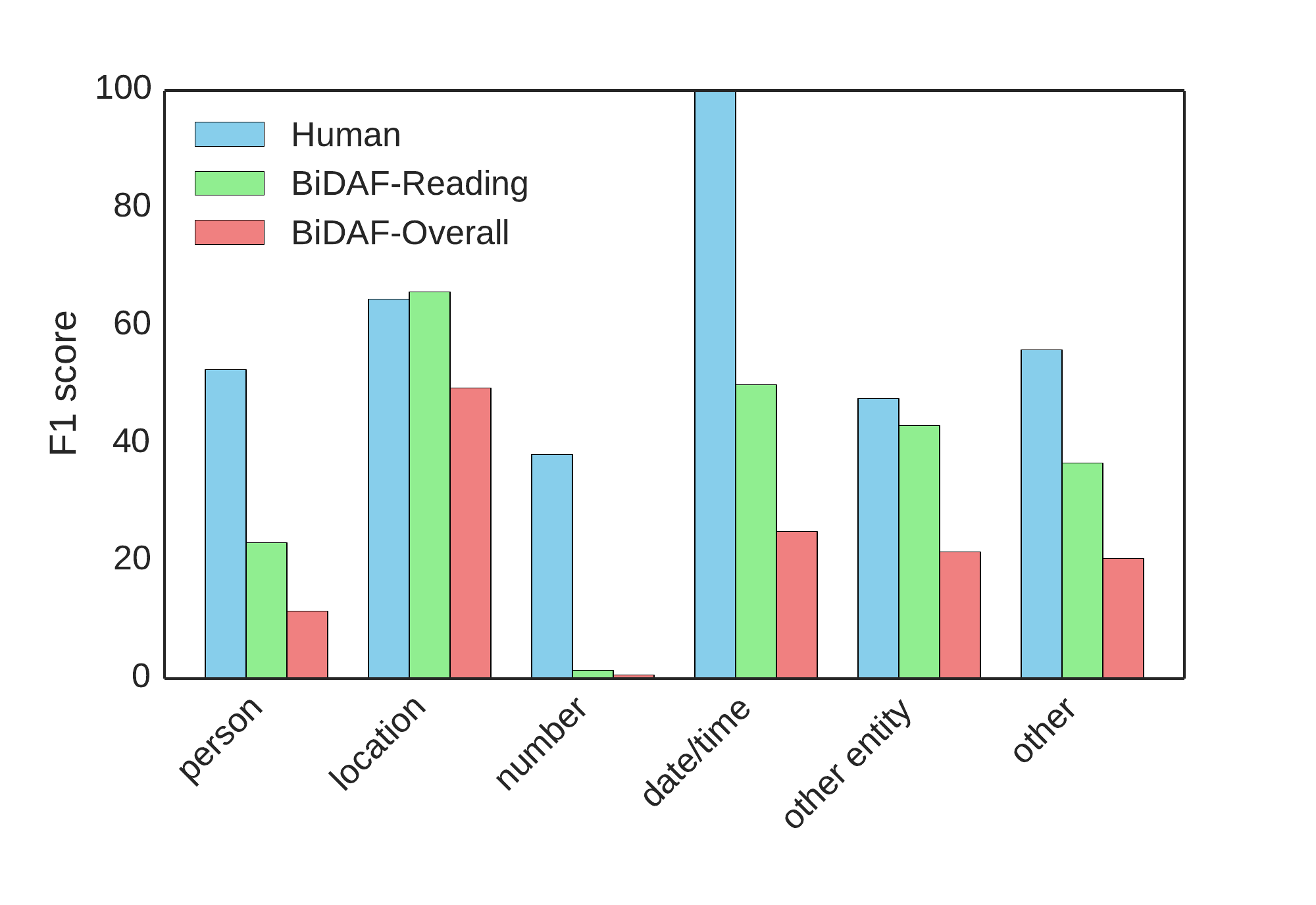}
        \caption{\textsc{Quasar-T} answer categories}
        \label{fig:tr_typ}
    \end{subfigure}
    \caption{\small Performance comparison of humans and the best performing baseline across the categories annotated for the development set.}\label{fig:annotperf}
\end{figure*}

Table~\ref{tab:descriptions} includes the definition of all the annotated relations for \textsc{Quasar-S}.

\section{Performance Analysis}
\label{app:performance}

Figure~\ref{fig:annotperf} shows a comparison of the human performance with the best performing baseline for each category of annotated questions. We see consistent differences between the two, except in the following cases. For \textsc{Quasar-S}, Bi-RNN performs comparably to humans for the \textit{developed-with} and \textit{runs-on} categories, but much worse in the \textit{has-component} and \textit{is-a} categories. For \textsc{Quasar-T}, BiDAF performs comparably to humans in the \textit{sports} category, but much worse in \textit{history \& religion} and \textit{language}, or when the answer type is a \textit{number} or \textit{date/time}.

\end{document}